\documentstyle[conference,named,lingmacros]{article} 

\begin{document}
\input{psfig}
\pagestyle{empty}
\date{}

\title{Discourse and Deliberation: Testing a
Collaborative Strategy \\
\begin{small} Coling 94, Kyoto \end{small}}
\author{Marilyn A. Walker \\
Mitsubishi Electric Research Laboratories\thanks{This research
was partially funded by ARO grant DAAL03-89-C0031PRI and DARPA grant
N00014-90-J-1863 at the University of Pennsylvania and by Hewlett
Packard, U.K.}    \\ 201 Broadway, Cambridge, Ma. 02139, USA
\\{\tt walker@merl.com} }

\maketitle
\bibliographystyle{named}

\begin{abstract}
\begin{quote}
A discourse strategy is a strategy for communicating with another
agent.  Designing effective dialogue systems requires designing agents
that can choose among discourse strategies. We claim that the design
of effective strategies must take cognitive factors into account,
propose a new method for testing the hypothesized factors,
and present experimental results on an effective strategy for
supporting deliberation. The proposed method of computational dialogue
simulation provides a new empirical basis for computational
linguistics.
\end{quote}
\end{abstract}


\section{Introduction}
\label{intro-sec}

A discourse strategy is a strategy for communicating with another
agent. Agents make strategy choices via decisions about when to talk,
when to let the other agent talk, what to say, and how to say it.  One
choice a conversational agent must make is whether an utterance should
include some relevant, but optional, information in what is
communicated. For example, consider \ex{1}:

\eenumsentence
{\item Let's walk along Walnut St.
 \item It's shorter.
\label{walnut-examp}
}

The speaker made a strategic {\bf choice} in \ex{0} to include
\ex{0}b since she could have simply said \ex{0}a. What
determines the speaker's choice?

Existing dialogue systems have two modes for dealing with optional
information: (1) include {\bf all} optional information that is not
already known to the hearer; (2) include {\bf no} optional information
\cite{MooreParis93}. But these modes are simply the extremes of
possibility and to my knowledge, no previous work has proposed any
principles for when to include optional information, or any way of
testing the proposed principles to see how they are affected by
the conversants and their processing abilities, by the
task, by the communication channel, or by the domain.

This paper presents a new experimental method for determining whether
a discourse strategy is effective and presents experimental results on
a strategy for supporting deliberation. The method is based on earlier
simulation work by Carletta and Pollack
\cite{Carletta92,PollackRinguette90}.  Section \ref{factor-sec}
outlines hypotheses about the factors that affect which strategies are
effective.  Section \ref{dw-sec} presents a new method for testing the
role of the hypothesized factors.  The experimental results in section
\ref{res-sec} show that effective strategies to support deliberation
are determined by both cognitive and task variables.

\section{Deliberation in Discourse}
\label{factor-sec}

\begin{figure*}[t]
\centerline{\psfig{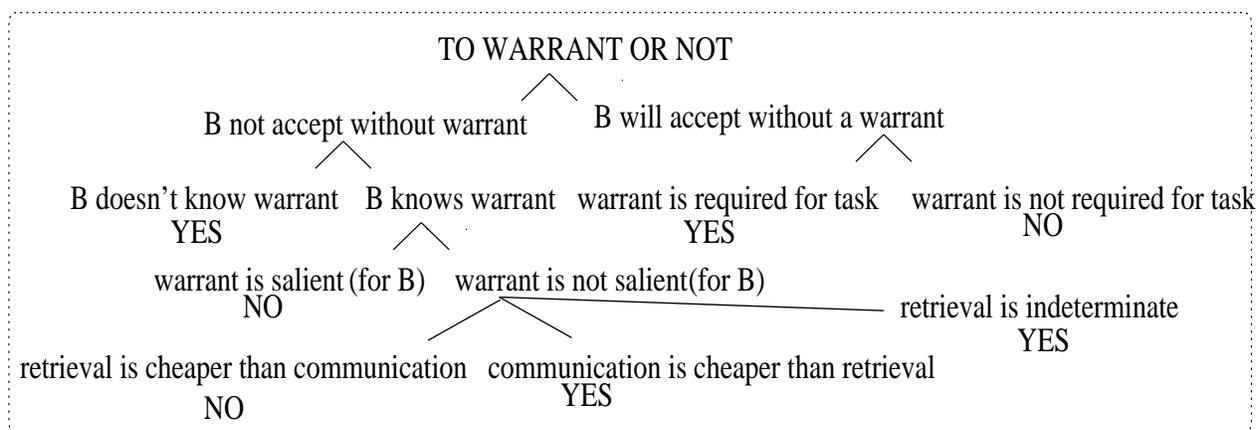}}
\caption{Potential Factors of Decision in whether to
use the Explicit-Warrant strategy}
\label{to-warr-fig}
\end{figure*}

Deliberation is the process by which an agent decides what to believe
and what to do \cite{Galliers91b,Doyle92}.  One strategy that
supports deliberation is the Explicit-Warrant strategy, as in
\ex{0}. The {\sc warrant} in \ex{0}b can be used by the hearer
in deliberating whether to {\sc accept} or {\sc reject} the speaker's
{\sc proposal} in \ex{0}a.\footnote{The relation between a {\sc
warrant} and the {\sc propose} communicative act is similar to the
{\sc motivation} relation of \cite{MooreParis93,MannThompson87}. A
{\sc warrant} is always optional; this is consistent with the RST
framework in which all satellites are optional information.}

An analysis of proposals in a corpus of 55 problem-solving dialogues
shows that communicating agents don't always include warrants in a
proposal, and suggest a number of hypotheses about which factors
affect their decision \cite{Walker93c,PHW82}.

Consider a situation in which an agent A wants an agent B to accept a
proposal P.  If B is a `helpful' agent (nonautonomous), B will accept
A's proposal without a warrant.  Alternatively, if B deliberates
whether to accept P, but B knows of no competing options, then P will
be the best option whether or not A tells B the warrant for P. Since a
warrant makes the dialogue longer, the Explicit-Warrant strategy might
be inefficient whenever either of these situations hold.

Now consider a situation where B is an autonomous agent
\cite{Galliers91b}. B always deliberates every proposal and
B probably knows of options which compete with proposal P.  Then B
cannot decide whether to accept P without a warrant. Supposedly agent
A should omit a warrant is if it is already believed by B, so that the
speaker in \ex{0} would not have said {\it It's shorter \/} if she
believed that the hearer knew that the Walnut St. route was shorter.
However, consider \ex{1}, said in discussing which Indian restaurant
to go to for lunch:

\eenumsentence
{\item Listen to Ramesh.
 \item He's Indian.
\label{ramesh-examp}
}

The warrant in \ex{0}b was included despite the fact that it was
common knowledge among the conversants. Its inclusion violates the
rule of {\it Don't tell people facts that they already
know}.\footnote{The {\sc warrant} having the desired effect of getting
the hearer to listen to Ramesh depends on the hearer previously
believing or coming to believe that Indians know of good Indian
restaurants \cite{WJ82}.} Clearly the rule does not hold. These
already-known warrants are a type of {\sc informationally redundant
utterance}, henceforth IRU, which are surprisingly frequent in
naturally-occurring dialogue \cite{Walker93c}.

A Warrant IRU such as that in \ex{0} suggests that B's cognitive
limitations may be a factor in what A chooses to say, so that even if
B {\bf knows} a warrant for adopting A's proposal, what is critical is
whether the warrant is {\bf salient} for B, i.e.  whether the warrant
is already accessible in B's working memory
\cite{Prince81,Baddeley86}.  If the warrant is not already salient,
then B must either infer or retrieve the warrant information or obtain
it from an external source in order to evaluate A's proposal.  Thus
A's strategy choice may depend on A's model of B's attentional state,
as well as the {\bf costs} of retrieval and inference as opposed to
communication. In other words, A may decide that it is easier to just
say the warrant rather than require B to infer or retrieve it.

Finally, the task determines whether there are penalties for leaving a
warrant implicit and relying on B to infer or retrieve it. Some tasks
require that two agents agree on the reasons for adopting a proposal,
e.g. in order to ensure robustness in situations of environmental
change. Other tasks, such as a management/union negotiation, only
require the agents to agree on the actions to be carried out and each
agent can have its own reasons for wanting those actions to be done
without affecting success in the task.

Figure \ref{to-warr-fig} summarizes these hypotheses by proposing a
hypothetical decision tree for an agent's choice of whether to use the
Explicit-Warrant strategy.  The choice is hypothesized to depend on
cognitive properties of B, e.g. what B knows, B's attentional state,
and B's processing capabilities, as well as properties of the task and
the communication channel.  To my knowledge, all previous work on
dialogue has simply assumed that an agent should never tell an agent
facts that the other agent already knows. The hypotheses in figure
\ref{to-warr-fig} seem completely plausible, but the relationship of
cognitive effort to dialogue behavior has never been explored.  Given
these hypotheses, what is required is a way to {\bf test} the
hypothesized relationship of task and cognitive factors to effective
discourse strategies.  Section
\ref{dw-sec} describes a new method for testing hypotheses about
effective discourse strategies in dialogue.

\section{Design-World}
\label{dw-sec}

Design-World is an experimental environment for testing the
relationship between discourse strategies, task parameters and agents'
cognitive capabilities, similar to the single agent TileWorld
simulation environment \cite{PollackRinguette90,HPC93}.  Design-World
agents can be parametrized as to discourse strategy, and the effects
of this strategy can be measured against a range of cognitive and task
parameters. This paper compares the Explicit-Warrant strategy to the
All-Implicit strategy as strategies for supporting deliberation. Other
strategies tested in Design-World are presented elsewhere
\cite{Walker93c,Walker94,RambowWalker94}.

\subsection{Design World Domain and Task}
\label{domain-sec}

\begin{figure}[ht]
\centerline{\psfig{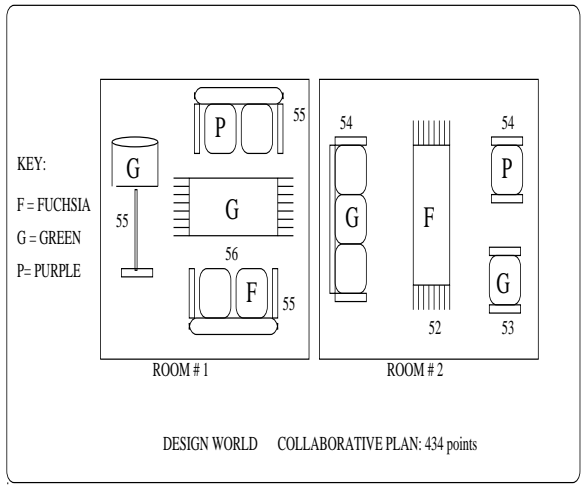}}
\caption{One Final State for Design-World Standard Task: Represents the
Collaborative Plan Achieved by the Dialogue, 434 points}
\label{final-state-fig}
\end{figure}

The Design-World task requires two agents to carry out a dialogue in
order to negotiate an agreement on the design of the floor plan of a
two room house \cite{WGR93}.  The {\sc design-house} plan requires the
agents to agree on how to {\sc design-room-1} and {\sc design-room-2}.
Both agents know what the {\sc design-house} plan requires and start
out with a set of furniture pieces that can be used to design each
room.

To negotiate an agreement, each agent carries out means-end reasoning
about the furniture pieces that they have that can be used in the
floor plan. Means-end reasoning generates {\sc options} -- these
options are the content of {\sc proposals} to the other agent to {\sc
put} a piece of furniture into one of the rooms.  Dialogue \ex{1}
illustrates agents' communication for part of designing room-1,
including both the artificial language that the agents communicate
with and a gloss generated from that language in {\it italics}:

\begin{small}
\enumsentence{
\item[1:]
BILL: {\it First, let's put the green rug in the study.} \\
(propose agent-bill agent-kim option-10: put-act (agent-bill green rug room-1))

\item[2:]
KIM: {\it Next, let's put the green lamp there.}

2:(propose agent-kim agent-bill option-33: put-act (agent-kim green lamp
room-1))
\item[3:]
BILL: {\it Then, let's put the green couch in the study.} \\
(propose agent-bill agent-kim option-45: put-act (agent-bill green couch
room-1))

\item[4:]
KIM: {\it No, instead let's put in the purple couch.} \\
(reject agent-kim agent-bill option-56: put-act (agent-kim purple couch
room-1))
}

\end{small}

On receiving a proposal, an agent deliberates whether to {\sc accept}
or {\sc reject} the proposal \cite{Doyle92}.  As potential
warrants to support deliberation, and to provide a way of
objectively evaluating agents' performance, each piece of furniture
has a score.  The score propositions for all the pieces of furniture
are stored in both agents' memories at the beginning of the dialogue.

Agents {\sc reject} a proposal if deliberation leads them to believe
that they know of a better option or if they believe the preconditions
for the proposal do not hold. The content of rejections is determined
by the {\sc collaborative planning principles}, abstracted from
analyzing four different types of problem solving dialogues
\cite{WW90,Walker94a}. For example, in \ex{0}-4 Kim rejects the
proposal in \ex{0}-3, and gives as her reason that option-56 is a
counter-proposal.

Proposals 1 and 2 are inferred to be implicitly {\sc accepted} because
they are not rejected \cite{WW90,Walker92a}.  If a proposal is {\sc
accepted}, either implicitly or explicitly, then the option that was
the content of the proposal becomes a mutual intention that
contributes to the final design plan \cite{Power84,Sidner92}. A
potential final design plan negotiated via a dialogue is shown in
figure \ref{final-state-fig}.

\subsection{Varying Discourse Strategies}
\label{strat-sec}

The Design-World experiments reported here compare the All-Implicit
strategy with the Explicit-Warrant strategy. Agents are parametrized
for different discourse strategies by placing different expansions of
discourse plans in their plan libraries. Discourse plans are plans for
{\sc proposal}, {\sc rejection}, {\sc acceptance}, {\sc
clarification}, {\sc opening} and {\sc closing}. The only variations
discussed here are variations in the expansions of {\sc proposals}.

The All-Implicit strategy is an expansion of a discourse plan to make
a {\sc proposal}, in which a {\sc proposal} decomposes trivially to
the communicative act of {\sc propose}.  In dialogue \ex{0}, both
Design-World agents communicate using the All-Implicit strategy, and
the proposals are shown in utterances 1, 2, and 3.  The All-Implicit
strategy never includes warrants in proposals, leaving it up to the
other agent to retrieve them from memory.

The Explicit-Warrant strategy expands the {\sc proposal} discourse act
to be a {\sc warrant} followed by a {\sc propose} utterance.  Since
agents already know the point values for pieces of furniture, warrants
are always IRUs in the experiments here. For example, \ex{1}-1 is a
{\sc warrant} for the proposal in \ex{1}-2: The names of agents who
use the Explicit-Warrant strategy are a numbered version of the string
``IEI'' to help the experimenter keep track of the simulation data
files; IEI stands for Implicit acceptance, Explicit warrant, Implicit
opening and closing.

\begin{small}
\enumsentence{

\item[1:]
IEI: {\bf Putting in the green rug is worth  56.} \\
(say agent-iei agent-iei2 bel-10: score  (option-10: put-act (agent-iei green
rug room-1) 56))

\item[2:]
IEI: {\it Then, let's put the green rug in the study.} \\
(propose agent-iei agent-iei2 option-10: put-act (agent-iei green rug room-1))

\item[3:]
IEI2: {\bf Putting in the green lamp is worth  55. } \\
(say agent-iei2 agent-iei bel-34: score  (option-33: put-act
(agent-iei2 green lamp room-1) 55)
 )

\item[4:]
IEI2: {\it Then, let's put the green lamp in the study.} \\
(propose agent-iei2 agent-iei option-33: put-act (agent-iei2 green lamp
room-1))
\label{exp-warr-dial}
}
\end{small}

The fact that the green rug is worth 56 points supports deliberation
about whether to adopt the intention of putting the green rug in the
study. The Explicit-Warrant strategy models naturally
occurring examples such as those in \ref{ramesh-examp} because the
points information used by the hearer to deliberate whether to accept
or reject the proposal is already mutually believed.

\subsection{Cognitive and Task Parameters}
\label{awm-sec}

Section \ref{factor-sec} introduced a range of factors motivated by the
corpus analysis that were hypothesized to determine when
Explicit-Warrant is an effective strategy. This section discusses how
Design-World supports the parametrization of these factors.


The agent architecture for deliberation and means-end reasoning is
based on the IRMA architecture, also used in the TileWorld simulation
environment \cite{PollackRinguette90}, with the addition of a
model of limited Attention/Working memory, AWM.
\cite{Walker93c} includes a fuller discussion of
the Design-World deliberation and means-end reasoning mechanism and
the underlying mechanisms assumed in collaborative planning.

We hypothesized that a warrant must be {\sc salient} for both agents
(as shown by example \ref{ramesh-examp}).  In Design-World, salience
is modeled by AWM model, adapted from \cite{Landauer75}. While the AWM
model is extremely simple, Landauer showed that it could be
parameterized to fit many empirical results on human memory and
learning \cite{Baddeley86}.  AWM consists of a three dimensional space
in which propositions acquired from perceiving the world are stored in
chronological sequence according to the location of a moving memory
pointer.  The sequence of memory loci used for storage constitutes a
random walk through memory with each loci a short distance from the
previous one.  If items are encountered multiple times, they are
stored multiple times
\cite{HintzmannBlock71}.

When an agent retrieves items from memory, search starts from the
current pointer location and spreads out in a spherical fashion.
Search is restricted to a particular search radius: radius is defined
in Hamming distance.  For example if the current memory pointer loci
is (0 0 0), the loci distance 1 away would be (0 1 0) (0 -1 0) (0 0 1)
(0 0 -1) (-1 0 0) (1 0 0). The actual locations are calculated modulo
the memory size. The limit on the search radius defines the capacity
of attention/working memory and hence defines which stored beliefs and
intentions are {\sc salient}.

The radius of the search sphere in the AWM model is used as the
parameter for Design-World agents' resource-bound on attentional
capacity.  In the experiments below, memory is 16x16x16 and the radius
parameter varies between 1 and 16, where AWM of 1 gives severely
attention limited agents and AWM of 16 means that everything an agent
knows is accessible.\footnote{The size of memory was determined as
adequate for producing the desired level of variation in the current
task across all the experimental variables, while still making it
possible to run a large number of simulations over night when agents
have access to all of their memory. In order to use the AWM model in a
different task, the experimenter might want to explore different sizes
for memory.} This parameter lets us distinguish between an agent's
{\bf ability} to access all the information stored in its memory, and
the effort involved in doing so.

The advantages of the AWM model is that it was shown to reproduce, in
simulation, many results on human memory and learning.  Because search
starts from the current pointer location, items that have been stored
most recently are more likely to be retrieved, predicting recency
effects \cite{Baddeley86}.  Because items that are stored in multiple
locations are more likely to be retrieved, the model predicts
frequency effects \cite{HintzmannBlock71}.  Because items are stored in
chronological sequence, the model produces natural associativity
effects \cite{Landauer75}.  Because deliberation and means-end reasoning can
only operate on salient beliefs, limited attention produces a
concomitant inferential limitation, i.e. if a belief is not salient it
cannot be used in deliberation or means-end-reasoning.  This means
that mistakes that agents make in their planning process have a
plausible cognitive basis. Agents can both fail to access a belief
that would allow them to produce an optimal plan, as well as make a
mistake in planning if a belief about how the world has changed as a
result of planning is not salient.  Depending on the preceding
discourse, and the agent's attentional capacity, the propositions that
an agent {\bf knows} may or may not be {\bf salient} when a proposal
is made.

Another hypothetical factor was the relative cost of retrieval and
communication.  AWM also gives us a way to measure the number of
retrievals from memory in terms of the number of locations searched to
find a proposition.  The amount of effort required for each retrieval
step is a parameter, as is the cost of each inference step and the
cost of each communicated message. These cost parameters support
modeling various cognitive architectures, e.g.  varying the cost of
retrieval models different assumptions about memory.  For example, if
retrieval is free then all items in working memory are instantly
accessible, as they would be if they were stored in registers with
fast parallel access. If AWM is set to 16, but retrieval isn't free,
the model approximates slow spreading activation that is
quite effortful, yet the agent still has the {\bf ability} to access
all of memory, given enough time. If AWM is set lower than 16 and
retrieval isn't free, then we model slow
spreading activation with a timeout when effort exceeds a certain
amount, so that an agent does not have the {\bf ability} to access all
of memory.

It does not make sense to fix absolute values for the retrieval,
inference and communication cost parameters in relation to human
processing. However, Design-World supports exploring issues about the
{\bf relative} costs of various processes.  These relative costs might
vary depending on the language that the agents are communicating with,
properties of the communication channel, how smart the agents are, how
much time they have, and what the demands of the task are
\cite{NormanBobrow75}. Below we vary the relative cost of
communication and retrieval.

Finally, we hypothesized that the Explicit-Warrant strategy may be
beneficial if the relationship between the warrant and the proposal
must be mutually believed. Thus the definition of success for the task
is a Design-World parameter: the Standard task does not require a
shared warrant, whereas the Zero NonMatching Beliefs task gives a zero
score to any negotiated plan without agreed-upon warrants.

\subsection{Evaluating Performance}
\label{eval-sec}


To evaluate {\sc performance}, we  compare the Explicit-Warrant
strategy with the All-Implicit strategy in situations where we
vary the task requirements, agents' attentional capacity, and the cost
of retrieval, inference and communication.  Evaluation of the
resulting {\sc design-house} plan is parametrized by (1) {\sc
commcost}: cost of sending a message; (2) {\sc infcost}: cost of
inference; and (3) {\sc retcost}: cost of retrieval from memory:

\begin{quote}
{\sc performance}
\begin{tabular}{ll}
    & $=$  Task Defined {\sc raw score} \\
    & -- ({\sc commcost} $\times$ number of messages)\\
    & -- ({\sc infcost} $\times$ number of inferences)\\
    & -- ({\sc retcost} $\times$ number of retrieval steps)
\end{tabular}
\end{quote}

{\sc raw score} is task specific: in the Standard task we simply
summarize the point values of the furniture pieces in each {\sc
put-act} in the final Design, while in the Zero NonMatching Beliefs
task, agents get no points for a plan unless they agree on the reasons
underlying each action that contributes to the plan.

The way {\sc performance} is defined reflects the fact that agents are
meant to collaborate on the task.  The costs that are deducted from
the {\sc raw score} are the costs for both agents' communication,
inference, and retrieval.  Thus {\sc performance} is a measure of {\sc
least collaborative effort} \cite{CS89,Brennan90}. Since the
parameters for cognitive effort are fixed while discourse strategy and
AWM settings are varied, we can directly test the benefits of
different discourse strategies under different assumptions about
cognitive effort and the cognitive demands of the task. This is
impossible to do with corpus analysis alone.

We simulate 100 dialogues at each parameter setting for each strategy.
Differences in performance distributions are evaluated for
significance over the 100 dialogues using the Kolmogorov-Smirnov (KS)
two sample test \cite{Siegel56}.

A strategy A is {\sc beneficial} as compared to a strategy B, for a
set of fixed parameter settings, if the difference in distributions
using the Kolmogorov-Smirnov two sample test is significant at p $<$
.05, in the positive direction, for two or more AWM settings.  A
strategy is {\sc detrimental} if the differences go in the negative
direction.  Strategies may be neither {\sc beneficial} or {\sc
detrimental}, as there may be no difference between two strategies.

\section{Results: Explicit Warrant}
\label{res-sec}

This section discusses the results of comparing the Explicit-Warrant
discourse strategy with the All-Implicit discourse strategy to
determine when each strategy is {\sc beneficial}. We test 4 factors
outlined in figure \ref{to-warr-fig}: when the warrant is salient or
not, when the warrant is required for the task or not, when the costs
of retrieval and communication vary, and when retrieval is
indeterminate.

Differences in performance between the Explicit-Warrant strategy and
the All-Implicit strategy are shown via a {\sc difference plot} such
as figure \ref{free-ret-iei-fig}. In figure \ref{free-ret-iei-fig}
performance differences are plotted on the Y-axis and AWM settings are
shown on the X-axis. If the plot is above the dotted line for 2 or
more AWM settings, then the Explicit-Warrant strategy may be {\sc
beneficial} depending on whether the differences are significant by
the KS test.  Each point represents the difference in the means of 100
runs of each strategy at a particular AWM setting. These plots
summarize the results of 1800 simulated dialogues: 100 for each AWM
setting for each strategy.

\subsubsection{Explicit Warrant reduces Retrievals}

\begin{figure}[htb]
\centerline{\psfig{figure=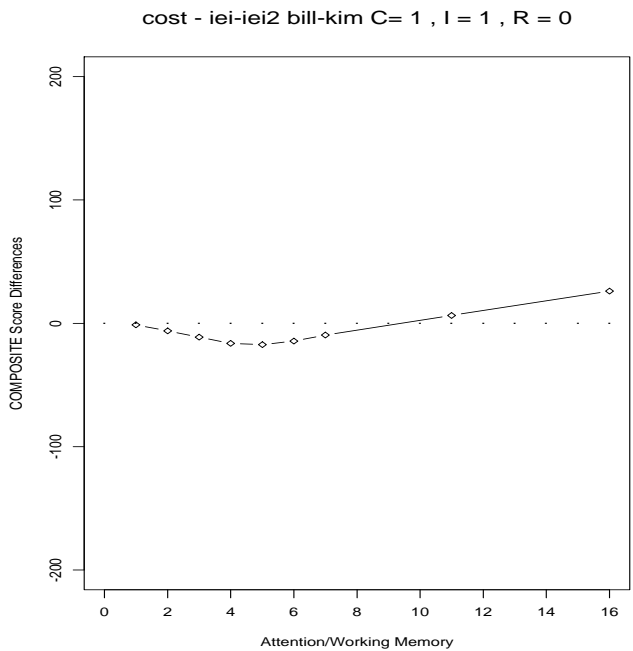,height=3.5in,width=3.5in}}
\caption{If Retrieval is Free, Explicit-Warrant is detrimental
at AWM of 3,4,5: Strategy 1 of two Explicit-Warrant agents and
strategy 2 of two All-Implicit agents: Task = Standard, commcost = 1, infcost =
1,
retcost = 0}
\label{free-ret-iei-fig}
\end{figure}

\begin{figure}[htb]
\centerline{\psfig{figure=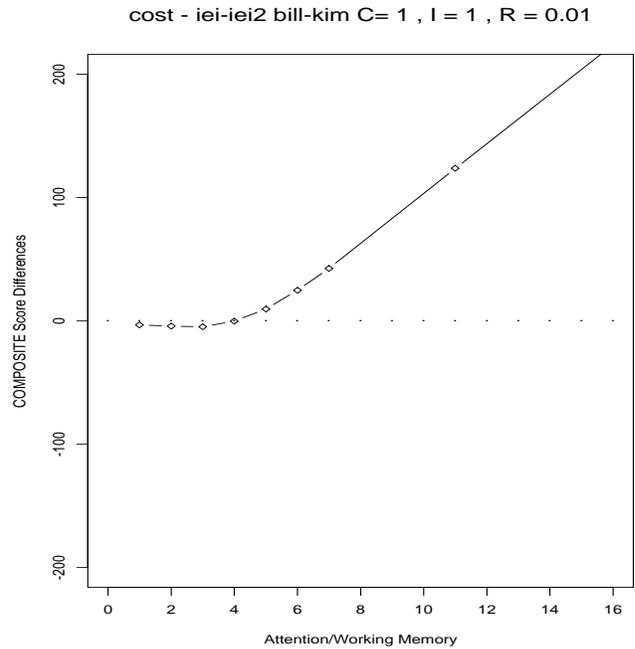,height=3.5in,width=3.5in}}
\caption{Retrieval costs: Strategy 1 is
two Explicit-Warrant agents and strategy 2 is two All-Implicit agents:
Task = Standard, commcost = 1, infcost = 1,
retcost = .01}
\label{ret-iei-fig}
\end{figure}


\begin{figure}[htb]
\centerline{\psfig{figure=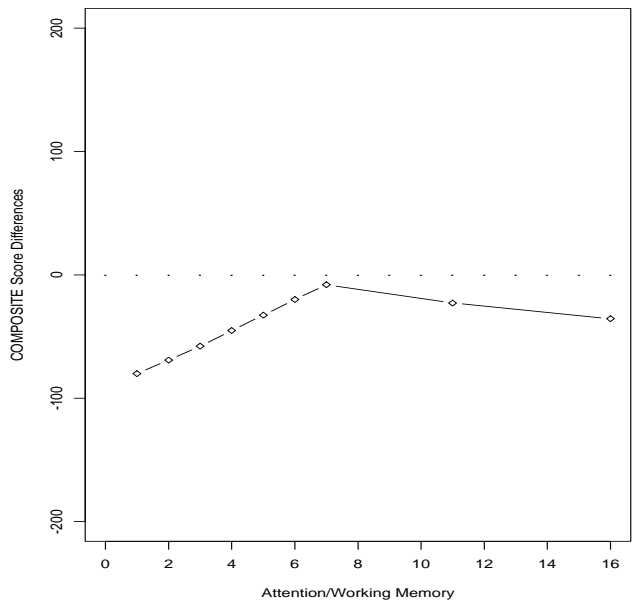,height=3.5in,width=3.5in}}
\caption{If Communication is Expensive: Communication costs can
dominate other costs in dialogues. Strategy 1 is
two Explicit-Warrant agents and strategy 2 is two All-Implicit agents:
Task = Standard, commcost = 10, infcost = 0,
retcost = 0}
\label{exp-comm-iei-fig}
\end{figure}


Dialogues in which one or both agents use the Explicit-Warrant
strategy are more efficient when retrieval has a cost.

Figure \ref{free-ret-iei-fig} shows that the Explicit-Warrant strategy
is {\sc detrimental} at AWM of 3,4,5 for the Standard task, in
comparison with the All-Implicit strategy, if retrieval from memory is
free (KS 3,4,5 $>$ .19, p $<$ .05).  This is because making the
warrant salient displaces information about other pieces of furniture
when agents are attention-limited. In the Standard task, agents aren't
required to share beliefs about the value of a proposal, so
remembering what pieces they have is more important than remembering
their value.

However, figure \ref{ret-iei-fig} shows that Explicit-Warrant is
beneficial when retrieval is one tenth the cost of communication and
inference.  By AWM values of 3, performance with Explicit-Warrant is
better than All-Implicit because the beliefs necessary for
deliberation are made salient with each proposal (KS for AWM of 3 and
above $>$ .23, p $<$ .01).  At AWM parameter settings of 16, where
agents have the ability to search all their beliefs for
warrants, the saving in processing time is substantial. Again at the
lowest AWM settings, the strategy is not beneficial because it
displaces information about other pieces from AWM.  However in figure
\ref{ret-iei-fig}, in contrast with figure \ref{free-ret-iei-fig},
retrieval has an associated cost. Thus the savings in retrieval
balance out with the loss of raw score so that the strategy is not
{\sc detrimental}. Other experiments show that even when the relative
cost of retrieval is .0001, that Explicit-Warrant is still beneficial
at AWM settings of 11 and 16 (KS for 11,16 $>$ .23 , p $<$ .01).


\subsubsection{Explicit Warrant is detrimental if Communication is Expensive}

If we change the relative costs of the different processes in the
situation, we change whether a strategy is beneficial.  Figure
\ref{exp-comm-iei-fig} shows that if communication cost is 10,
and inference and retrieval are free, then the Explicit-Warrant
strategy is {\sc detrimental} (KS for AWM 1 to 5 $>$ .23, p$<$ .01).
This is because the Explicit-Warrant strategy increases the number of
utterances required to perform the task; it doubles the number of
messages in every proposal. If communication is expensive compared to
retrieval, communication cost can dominate the other benefits.

\subsubsection{Explicit Warrant Achieves a High Level of Agreement}

If we change the definition of success in the task, we change whether
a strategy is beneficial.  When the task is Zero-Nonmatching-Beliefs,
the Explicit-Warrant strategy is beneficial even if retrieval is free
(KS $>$ .23 for AWM from 2 to 11, p $<$ .01) The warrant information
that is redundantly provided is exactly the information that is needed
in order to achieve matching beliefs about the warrants for intended
actions. The strategy virtually guarantees that the agents will agree
on the reasons for carrying out a particular course of action. The
fact that retrieval is indeterminate produces this effect; a similar
result is obtained when warrants are required and retrieval costs
something.

\begin{figure}[htb]
\centerline{\psfig{figure=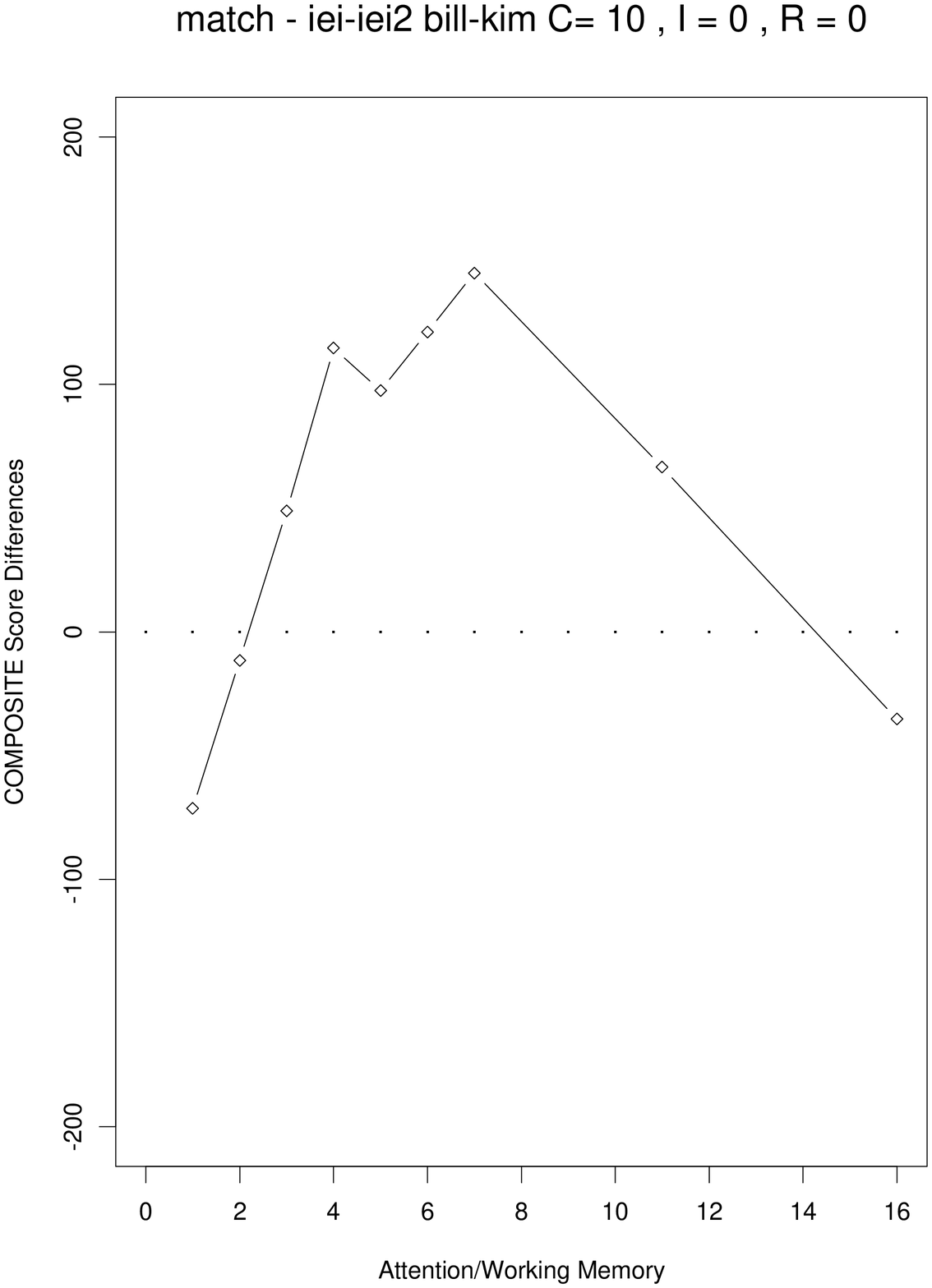,height=3.5in,width=3.5in}}
\caption{Explicit-Warrant is STILL beneficial: Strategy 1 is
two Explicit-Warrant agents and strategy 2 is two All-Implicit agents:
Task =  Zero-Nonmatching-Beliefs, commcost = 10, infcost = 0,
retcost = 0}
\label{iei-nmb-high-cost-fig}
\end{figure}

To my great surprise, the beneficial effect of Explicit-Warrant for
the Zero-NonMatching-Beliefs task is so robust that even if
communication cost is 10 and retrieval and inference are free,
Explicit-Warrant is better than All-Implicit at AWM of 3 $\ldots$ 11
(KS $>$ .23, p $<$ .01).  See figure \ref{iei-nmb-high-cost-fig}. In
other words, even when every extra {\sc warrant} message incurs a
penalty of 10 points, if the task is Zero-NonMatching-Beliefs, agents
using Explicit-Warrant do better. Contrast figure
\ref{iei-nmb-high-cost-fig} with the Standard task and same cost
parameters in \ref{exp-comm-iei-fig}.

These result suggests that including warrants is highly effective when
agents must agree on a specific warrant, if they are attention-limited
to any extent.



\section{Conclusion}
\label{conc-sec}

This paper has discussed an instance of a general problem in the
design of conversational agents: when to include optional information.
We presented and tested a number of hypotheses about the factors that
contribute to the decision of when to include a warrant in a proposal.
We showed that warrants are useful when the task requires agreement on
the warrant, when the warrant is not currently salient, when retrieval
of the warrant is indeterminate, or when retrieval has some associated
cost, and that warrants hinder performance if communication is costly
and if the warrant can displace information that is needed to complete
the task, e.g. when AWM is very limited and warrants are not required
to be shared.

The method used here is a new experimental methodology for
computational linguistics that supports testing hypotheses about
beneficial discourse strategies \cite{Carletta92,PollackRinguette90}.
The Design-World environment is based on a cognitive model of limited
attention and supports experiments on the interaction of discourse
strategies with agents' cognitive limitations. The use of the method
and the focus of this work are novel: previous work has focused on
determining underlying mechanisms for cooperative strategies rather
than on investigating when a strategy is effective.

To my knowledge, no previous work on dialogue has ever argued that
conversational agents' resource limits are a major factor in
determining effective conversational strategies in collaboration.  The
results presented here suggest that cooperative strategies cannot be
defined in the abstract, but cooperation arises from the {\bf
interaction} of two agents in dialogue. If one agent has limited
working memory, then the other agent can make the dialogue go more
smoothly by adopting a strategy that makes deliberative premises
salient. In other words, strategies are cooperative {\bf for} certain
conversational partners, under particular task definitions, for
particular communication situations.

Here we compared two discourse strategies: All-Implicit and
Explicit-Warrant. Explicit-Warrant is a type of discourse strategy
called an Attention strategy in \cite{Walker93c} because its main
function is to manipulate agents' attentional state.
Elsewhere we show that (1) some IRU strategies are only beneficial
when inferential complexity is higher than in the Standard Task
\cite{RambowWalker94,Walker94}; (2) IRUs that make inferences explicit can help
inference limited agents perform as well as logically omniscient ones
\cite{Walker93c}.

Although much work remains to be done, there is reason to believe that
these results are domain independent.  The simplicity of the
Design-World task means that its structure is a subcomponent of many
other tasks. The model of limited resources is cognitively based, but
the cost parameters support modeling different agent architectures,
and we explored the effects of different cost parameters. The
Explicit-Warrant strategy is based on simple relationships between
different facts which we would expect to occur in any domain, i.e. the
fact that some belief can be used as a {\sc warrant} for accepting a
proposal should occur in almost any task.  Future work should extend
these results, showing that a `cooperative strategy' need not always
be `cooperative', and investigate additional factors that determine
when strategies are effective.

\end{document}